\documentclass{article}
\usepackage{PRIMEarxiv}
\usepackage{natbib}
\usepackage[utf8]{inputenc} % allow utf-8 input
\usepackage[T1]{fontenc}    % use 8-bit T1 fonts
\usepackage{hyperref}       % hyperlinks
\usepackage{url}            % simple URL typesetting
\usepackage{booktabs}       % professional-quality tables
\usepackage{amsfonts}       % blackboard math symbols
\usepackage{nicefrac}       % compact symbols for 1/2, etc.
\usepackage{microtype}      % microtypography
\usepackage{lipsum}
\usepackage{fancyhdr}       % header
\usepackage{graphicx}       % graphics
\usepackage{amsmath}
\graphicspath{{media/}}     % organize your images and other figures under media/ folder
\usepackage{colortbl} % Load the package for coloring rows
\usepackage{xcolor} % Load the package for defining colors
\definecolor{LightBlue}{rgb}{0.68, 0.85, 0.9}
\definecolor{LightGreen}{rgb}{0.56, 0.93, 0.56}

\title{Predicting Performance of Object Detection Models in Electron Microscopy Using Random Forests
\thanks{\textit{\underline{Citation}}: 
\textbf{Authors. Title. Pages.... DOI:000000/11111.}} 
}
\author{
  Ni Li, Ryan Jacobs, Vidit Agrawal, Dane Morgan \\
  Affiliation \\
  University of Wisconsin-Madison \\
  Madison, WI\\
  \texttt{\{nli72, rjacobs3, vidit.agrawal, ddmorgan\}@wisc.edu} \\
   \And
  Matthew Lynch,  Kevin Field \\
  Affiliation \\
  University of Michigan \\
  Ann Arbor, MI\\
  \texttt{\{mjdl, kgfield\}@umich.edu} \\
}
\begin{document}
\maketitle
\begin{abstract}
Quantifying prediction uncertainty when applying object detection models to new, unlabeled datasets is critical in applied machine learning. This study introduces an approach to estimate the performance of deep learning-based object detection models for quantifying defects in transmission electron microscopy (TEM) images, focusing on detecting irradiation-induced cavities in TEM images of metal alloys. We developed a random forest regression model that predicts the object detection F1 score, a statistical metric used to evaluate the ability to accurately locate and classify objects of interest. The random forest model uses features extracted from the predictions of the object detection model whose uncertainty is being quantified, enabling fast prediction on new, unlabeled images. The mean absolute error (MAE) for predicting F1 of the trained model on test data is 0.09, and the $R^2$ score is 0.77, indicating there is a significant correlation between the random forest regression model predicted and true defect detection F1 scores. The approach is shown to be robust across three distinct TEM image datasets with varying imaging and material domains. Our approach enables users to estimate the reliability of a defect detection and segmentation model predictions and assess the applicability of the model to their specific datasets, providing valuable information about possible domain shifts and whether the model needs to be fine-tuned or trained on additional data to be maximally effective for the desired use case.
\end{abstract}

% keywords can be removed

\section{Introduction}
Electron microscopy (EM) techniques are among the most effective tools to characterize the structure of materials. Among the EM techniques, transmission electron microscopy (TEM) has been widely used to study defects in materials owing to its ability to visualize individual defects at the atomic to nanometer scale\cite{LeeACS}. The identification and annotation of objects of interest in TEM images (\textit{e.g.}, atomic vacancies, dislocation loops, cavities, etc.) has traditionally been accomplished manually by domain-expert scientists \cite{10.1017/S1551929522001286,JACOBS2022111527}. However, these manual methods suffer from human-related inconsistencies (\textit{e.g.}, bias toward identifying certain features and excluding others) and are not automatically scalable, especially given the modern EM instruments' capability to generate large volumes of complex data. The drawbacks associated with manual labeling necessitate an automated approach, where machine learning (ML), particularly deep learning (DL), has emerged as a viable solution.

In recent years, DL has significantly advanced the fields of computer vision and image processing. Specifically, convolutional neural networks (CNNs), due to their ability to efficiently and accurately identify relevant features in images, have been transformative and widely applied to identify objects within images with high accuracy. Advanced CNNs like ResNet50, VGG16 and U-net\cite{ronneberger2015unetconvolutionalnetworksbiomedical} have become foundational in object detection frameworks, such as the Faster Regional Convolutional Neural network (Faster R-CNN)\cite{ren2016faster}, Mask R-CNN\cite{he2018mask} and YOLO (you only look once)\cite{redmon2016look}. These and related object detection frameworks have recently gained significant traction in materials research, and have been employed to detect features such as void defects, dislocation loops and nanoparticles\cite{RN987,RN985,CHEN2023112073,RN988,Jacobs2023,RN984,RN985,RN986,RN989,RN990,Holm2020,DENNLER2021103069,OKTAY2019113}. Although not directly related to this work, models based on fully convolutional networks (FCNs) have also been employed to locate individual atoms in EM images\cite{LeeACS,doi:10.1021/acsnano.7b07504,Lin2021}.

Overall, object detection models have achieved human domain-expert level performance (with dramatically faster prediction times) for characterizing the numbers, shapes and sizes of various defect types in EM images for numerous types of materials\cite{JACOBS2022111527}. However, it has been pointed out that the performance of the object detection models vary with the overall quality of EM images, the size and visual quality of individual objects to be identified, and the selection of training and testing data used to train the object detection model. For example, Jacobs et al.\cite{RN985} found that the performance of a Mask R-CNN model for detecting defects in TEM images was affected by the similarity between images comprising the training and testing dataset, where it was found that testing images from a different data source, material type or imaging condition than was included in the training data resulted in significantly degraded model performance. Wei et al. (2022)\cite{10.1093/micmic/ozac043} demonstrated the significant impact of STEM image quality (such as resolution and contrast) and the similarity to the training data on the performance of FCN-based models. It has also been observed that the robustness of neural networks varies with EM images taken with different experimental parameters, such as magnification and electron dosage\cite{10.1093/micmic/ozae001}. Finally, Jacobs et al.\cite{Jacobs2023} found that a Mask R-CNN model to characterize cavities in TEM images of irradiated metal alloys had difficulty in detecting small cavities (i.e., those less than a few percent of the image dimension), and Bruno et al. found that human labelers, even domain-expert ones, will introduce biases into their ground-truth labeling when attempting to label objects that are small or visually ambiguous.\cite{10.1093/micmic/ozad067.767} The examples provided above leveraged significant scientific-domain expertise to identify when certain data was likely to fall inside or outside the applicability domain of the trained object detection model. Such information is not always readily available or practical to obtain, and having some uncertainty quantification of object detection model predictions would be highly beneficial for application of object detection models for EM image characterization.

\begin{figure*}[ht]
 \centering
 \includegraphics[width=0.95\textwidth]{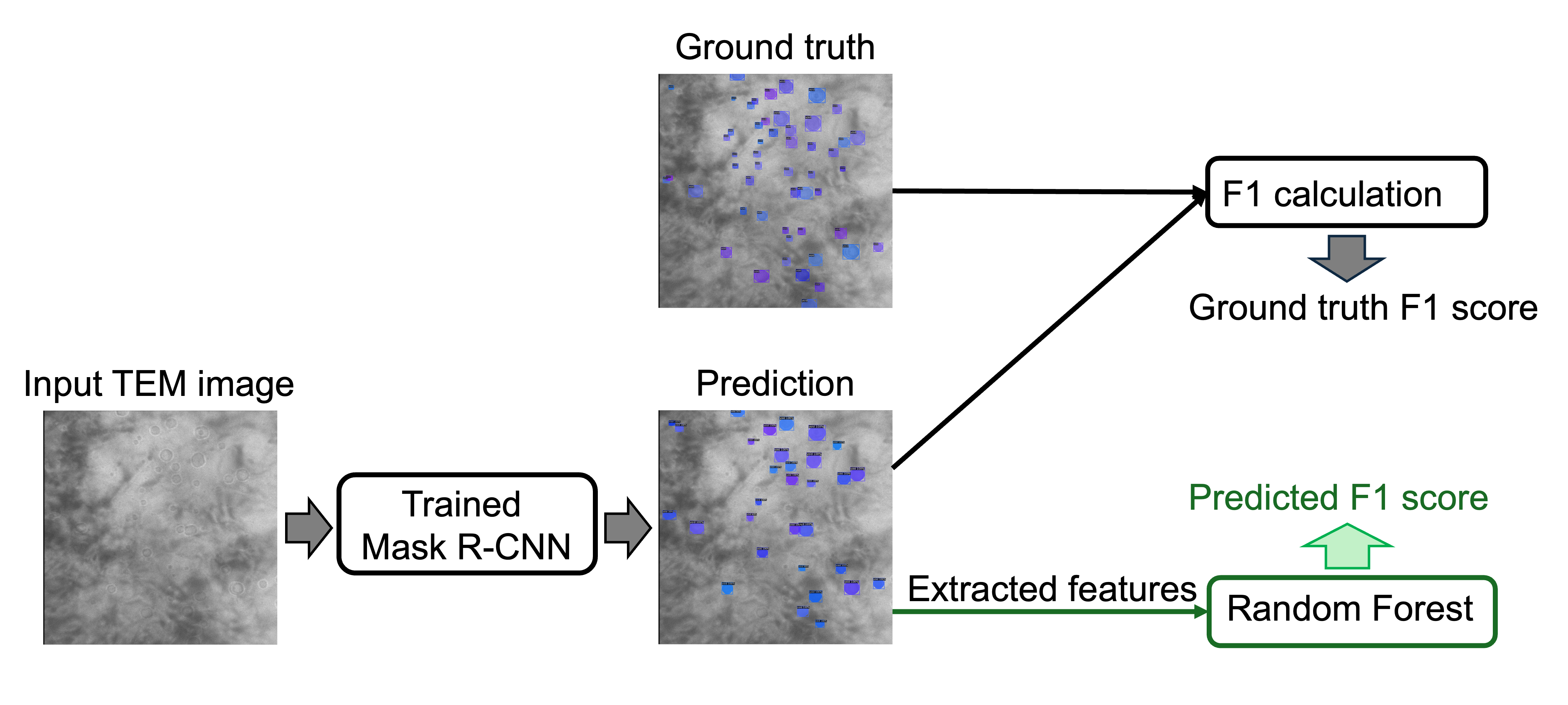}
\caption{Workflow diagram illustrating the process of estimating the defect detection F1 score using a trained Mask R-CNN model. The procedure includes using the trained Mask R-CNN to identify defects in TEM images, extracting key features from the predicted defects, and utilizing a Random Forest Regression model to predict the F1 score, thereby estimating the performance without the need for ground truth labels.}
 \label{fig:fig1}
\end{figure*}

The success of deep neural networks in the field of computer vision is dependent on the presumption that the data used for training and testing are drawn from the same distribution\cite{BenDavid2010ATO}\cite{7410480}. The decline in performance when applied to data that deviates from the distribution seen during training is commonly referred to as the out-of-distribution (OOD) problem\cite{hendrycks2017a}\cite{Yang2024}. In computer vision, OOD detection has traditionally been framed as a classification task to distinguish between OOD and in-distribution samples\cite{Yang2024}\cite{vinyals2017starcraft}. Commonly-used image benchmarks, like CIFAR and ImageNet, consist predominantly of visually distinct common objects (\textit{e.g.}, pictures of individual animals, furniture, food, people, etc.). In EM imaging, however, object variations and distinctions are typically much less obvious, where even different domain-expert labelers will show marked differences in apparent ground truth labeling\cite{lynch2024acceleratingdomainawareelectronmicroscopy}\cite{10.1093/micmic/ozad067.767}. Therefore, the approach to treat OOD detection in EM images as a binary classification problem is not feasible due to minute but distinct varying imaging domains, nuanced labeling, and complex evaluation criteria. In this work, our focus is on developing an approach that estimates the likely accuracy of a DL defect detection model for a given image so that the user can decide how they wish to use the predictions from that image. 

There are two main approaches to address EM image-based DL model uncertainty, depending on its origin and the objective. In automated experimentation, data distribution may experience OOD drift due to the acquisition of new data, leading to decreased model performance\cite{Kalinin2023}. The goal in such scenarios is to enhance model performance, with current methods focusing on the iterative training of ML models to enable adaptive learning as the underlying data used in training is updated\cite{Ghosh2021}. This is an exciting approach, but involves a significant effort associated with obtaining consistently labeled data and retraining models to address issues. Another approach concerns the treatment of outlier EM images, such as those that are empty or exhibit a low signal-to-noise ratio, and therefore lack valuable information and should be discarded. Here, the objective is simply to flag and reject these outlier images, not use them for retraining, and thereby ensure the integrity of the data used for analysis\cite{Sorzano2014OutlierDF}. However, determining outliers can be challenging since model performance depends on many factors. We take an approach similar to this second outlier approach, although we provide a continuous prediction of quality (i.e., predicted F1 score) rather than just a classification of in-distribution or OOD. 

In this work we develop and validate a performance estimation framework capable of predicting how well a trained Mask R-CNN model is expected to locate and classify objects when applied to new TEM images. Although we focus on one model type and just cavity defects in irradiated metal alloys, we expect the overall approach to be useful for quantifying the performance of many object detection models trained on many different types of objects and images. Crucially, our trained random forest model can be applied to images for which no labeled ground truth data is available, providing insight for the expected performance of the object detection model on new, unseen data. Figure \ref{fig:fig1} illustrates the workflow of the performance evaluation procedure without ground truth labels. Rather than simplifying the problem to a binary classification of data to in-distribution or OOD, we have developed a methodology that predicts the defect detection F1 score as a metric for a quantitative evaluation of model performance. We have trained a random forest regression model to learn the relationship between selected features derived from the Mask R-CNN model output (the bounding boxes and associated confidence scores) and the object detection F1 score. By processing new images through a pre-trained Mask R-CNN model, one can subsequently employ our random forest regression model to estimate the defect detection F1 score. This predictive capability allows users of our Mask R-CNN model to estimate the reliability of their results and determine the suitability of the model to their specific datasets. Our framework is particularly useful in applying trained defect detection models on new images where image quality and characteristics may be different from the training dataset, \textit{e.g}., due to domain shift and/or just poor image quality. This work also opens new avenues for the robust application of machine learning models in materials science, where understanding and quantifying uncertainty is crucial for advancing experimental and analytical techniques.

\section{Data and Methods}
\label{sec:headings}
\subsection{Data acquisition}
\begin{figure*}[ht]
 \centering
 \includegraphics[width=0.95\textwidth]{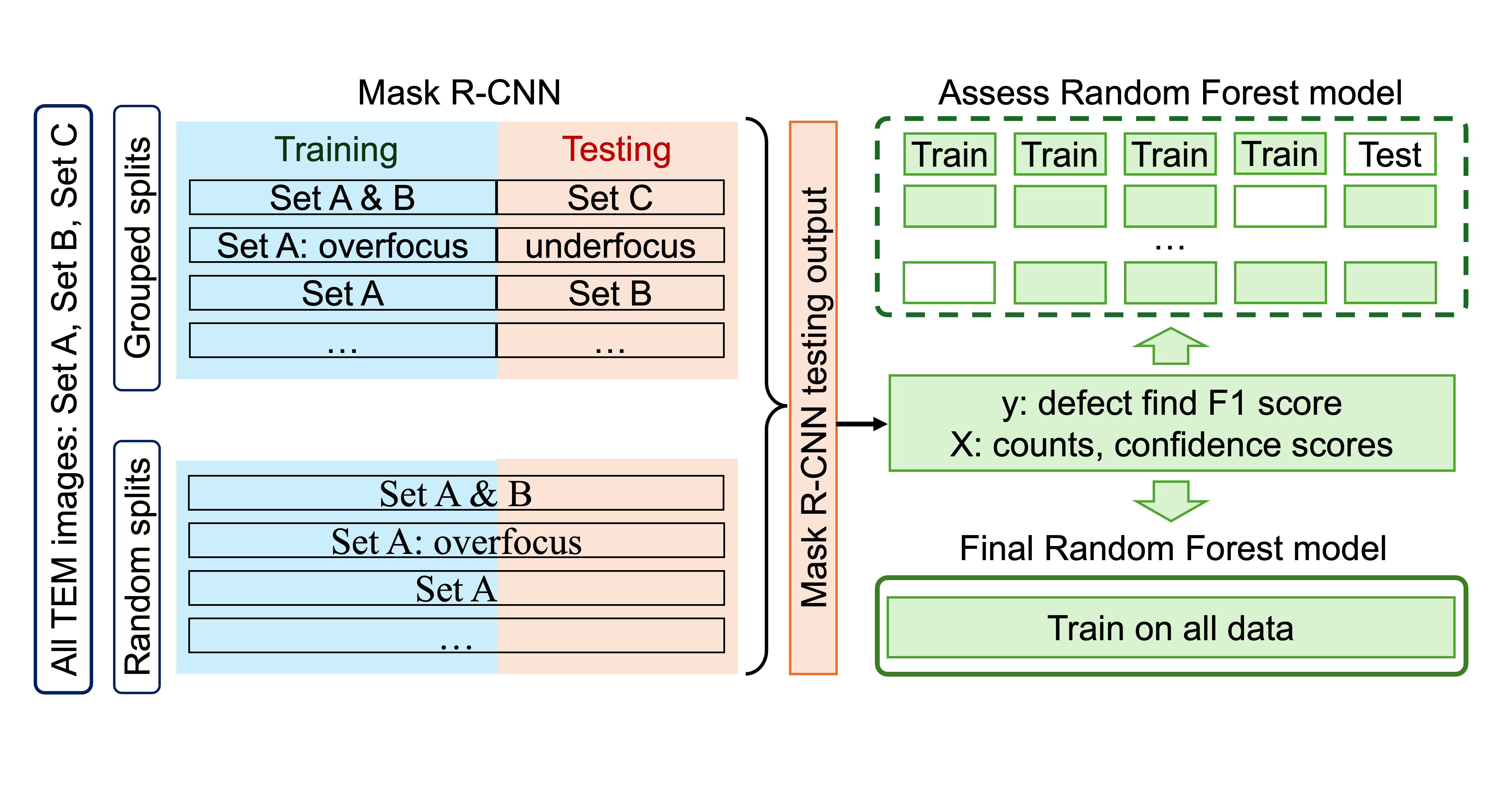}
\caption{Data Generation and Utilization Workflow. This flowchart illustrates the sequential steps undertaken in our study, starting from the collection of TEM images, through the training and evaluation of the Mask R-CNN model, to the feature extraction and final F1 score prediction using Random Forest regression. The data is distinctly categorized for Mask R-CNN training and testing, followed by a five-fold cross-validation scheme applied in the Random Forest training phase, highlighting the two experimental setups: consistent source (random splits) and varied source (grouped splits) between training and testing datasets.}
 \label{fig:dataflow}
 \end{figure*}
 
The three datasets: Set A, Set B and Set C used in our study comprise TEM images of cavities in metal alloys which have undergone neutron or ion irradiation. TEM images from the three different sets vary in the material composition and/or structure, irradiation condition, TEM instrument used, TEM imaging conditions, and ground truth labels. Previous utilization of Set A and Set B is documented in the work of Jacobs et al.\cite{Jacobs2023} and Lynch et al. \cite{lynch2024acceleratingdomainawareelectronmicroscopy}, where detailed descriptions of these datasets are available. Images in Set A were taken of steel alloys with various compositions irradiated by neutrons or ions, obtained at the Nuclear Oriented Materials \& Examination (NOME) Laboratory at the University of Michigan. Set B contains images of irradiated X-750 alloy with helium bubbles generated by Canadian Nuclear Laboratories (CNL)\cite{ANDERSON2020113068}. The TEM images within Set C originate from samples of Fe and Fe-10Cr alloys irradiated by Kr and He ions at the Intermediate-Voltage Electron Microscopy (IVEM)-Tandem facility at Argonne National Laboratory (ANL)\cite{CHEN2023112073}. The objects targeted for detection are cavities (sometimes also called voids or bubbles) in TEM images and typically exhibit circular or faceted shapes. Notably, all three datasets include images that have imaging conditions that are either underfocused or overfocused to form Frensel contrast in the images. The voids in TEM images with Fresnel contrast appear with bright boundary pixels when captured in the overfocus mode and with dark boundary pixels in the underfocus condition. For the purposes of our analysis, Set A was subdivided into two subgroups - Set A: underfocus and Set A: overfocus, which is to facilitate data partition in the training and testing phases of the Mask R-CNN model.

The data generation and utilization workflow shown in Figure \ref{fig:dataflow} begins with a comprehensive collection of the three sets of TEM images described above, from which a subset is used for the training of the Mask R-CNN model, and a distinct subset of the TEM images is deployed to test the performance of the trained Mask R-CNN. As shown in Table \ref{table:splits}, the data splits used to train and test the Mask R-CNN model includes two types of splits: one where the training and testing datasets are sourced from the same subset (these are random splits so the test data is likely to be in the same distribution as the training data), and another where the testing data are sourced from a different subset than the training data (these are splits based on distinct subsets with known significant differences so the test data is likely outside the distribution of training data). We will refer to these carefully designed distinct subsets as "grouped" subsets to reflect the distinct nature of their grouping. The method of determining what is in each grouped subset is based on either (i) data coming from different origins, \textit{e.g.}, Set A vs. Set B, and thus represent different materials, irradiation conditions, and TEM instruments, or (ii) data coming from different imaging modes, where here the main difference in imaging mode is overfocus vs. underfocus conditions. For each case, a Mask R-CNN model was trained on the training dataset and then was applied to detect cavities in the test images. The resulting bounding boxes and their confidence scores on the test images were used as a basis for creating features to train the random forest model to predict the object detection F1 score, discussed more in Sec. \ref{subsec:rfr}.

\begin{table}[h!]
\centering
\begin{tabular}{c|c|p{2 cm}|p{2 cm}}
\hline
\textbf{Split} & \textbf{Notation} &\textbf{Mask R-CNN trained on} & \textbf{Mask R-CNN tested on} \\ \hline
\rowcolor{LightBlue}

 1 &A\&B\_C & A\&B & C \\ \hline
 \rowcolor{LightBlue}
 2 &A: over\_under & A: overfocus & A: underfocus \\ \hline
 \rowcolor{LightBlue}
 3 &A: under\_over & A: underfocus & A: overfocus \\
        \hline
         \rowcolor{LightBlue}

        4 &B\_A & B & A \\
        \hline
         \rowcolor{LightBlue}

        5 &A\_B & A & B \\
        \hline
         \rowcolor{LightGreen}

        6 &A\&B\_A\&B & A\&B & A\&B \\
        \hline
        \rowcolor{LightGreen}
        7 &A\_A & A & A \\
        \hline
        \rowcolor{LightGreen}
        8 &B\_B & B & B \\
        \hline
        \rowcolor{LightGreen}
        9 &A: under\_under & A: underfocus & A: underfocus \\
        \hline
        \rowcolor{LightGreen}
        10 &A: over\_over & A: overfocus & A: overfocus \\
        \hline

%        \bottomrule
\end{tabular}
\caption{List of splits and how they were obtained.}
\label{table:splits}
\end{table}

\subsection{Mask R-CNN model and assessment}
\label{subsec:mask}

The structure and implementation of the Mask R-CNN model used in this work is the same as that used in the work of Jacobs et al.\cite{Jacobs2023} (Detectron2 implementation with PyTorch backend), and more details about model training and hyperparameters can be found in that study. The Intersection over Union (IoU) is a measure used to quantify the overlap between the bounding boxes of two objects. In this study, following Jacobs et al.,\cite{Jacobs2023} an IoU threshold of 0.1 indicates that a prediction is considered a true positive if at least 10\% of the predicted area overlaps with the ground truth. The confidence score is a likelihood measure that the Mask R-CNN model region proposal contains an object of interest. Here, a confidence threshold of 0.1 was adopted following previous work\cite{Jacobs2023}. The F1-score is a measure of the defect detection model's performance on test images and serves as the learning target (y) for the regression model. The F1 score of each image was calculated by comparing the ground truth manual annotations and predictions made by the Mask R-CNN model using Eq (1): 
\begin{equation}
\text{F1-score} = \frac{2}{\text{precision}^{-1} + \text{recall}^{-1}}=\frac{2\times\text{TP}}{2\times\text{TP}+\text{FP}+\text{FN}}
\end{equation}

Where true positives (TP) denote the number of correctly detected defects, false positives (FP) denote the number of predicted defects which are not defects in the ground truth images, and false negatives (FN) denote the number of defects labeled in the ground truth images but not predicted by the Mask R-CNN model. A high F1 score (\textit{e.g.}, typically 0.7 or higher, though this value depends on the application) indicates good performance.
 
\subsection{Random forest model and assessment}
\label{subsec:rfr}

Random forest is one of the most widely used ML methods in materials science\cite{JAIN2024101189}\cite{annurev:/content/journals/10.1146/annurev-matsci-070218-010015} due to its robustness, ease of use, and ability to handle nonlinear relationships between features and the target variable. The random forest model works by constructing multiple decision trees during training, each of which is fit to a separate bootstrapped sample of the training data, and outputting the mean prediction of the individual trees. This ensemble method helps improve accuracy and minimize over-fitting. The performance of a random forest model may vary with the number of trees in the forest. In our case, we utilized 100 trees to balance overall model complexity and performance.

As shown in Fig. \ref{fig:dataflow}, the performance of the random forest regression model was assessed using either random five fold cross-validation (random splits) or leave-out-group cross-validation (grouped splits). The final model used for deployment was fit on all of the data together. The performance of the trained model on each test dataset was evaluated using five well-established evaluation metrics. The obtained evaluation metrics were averaged over five test folds to reflect the overall performance of the model. Apart from the three widely used metrics: the coefficient of determination ($R^2$), the root mean square error (RMSE), and the mean absolute error (MAE), the normalized RMSE (NRMSE) and normalized MAE (NMAE) are also employed. NRMSE normalizes the RMSE by the standard deviation of the ground truth F1 scores in the test set being considered, while NMAE normalizes the MAE relative to the mean of the ground truth F1 scores in the test set being considered.

\subsection{Feature engineering}

The output from the Mask R-CNN model for each image is a list of detected cavity bounding boxes and the corresponding confidence scores. To featurize the random forest model, we pursued an approach that selected optimal features from a long feature candidate list derived from different quantification of the distributions of the detected cavity sizes, cavity counts and cavity confidence scores. Our initial feature set contained the following candidate features, calculated for each image: (1-9) The confidence scores (ranges from 0.1 to 1) were segmented into 9 distinct bins, and the counts of scores within each bin were calculated and divided by the total number of detected defects; (10) the area ratio (defined as the combined area of all detected cavities relative to the total image area), (11) the average confidence score, (12) standard deviation of confidence score, (13) average fractional detected defect bounding box size, (14) standard deviation of fractional detected defect size, (15) the average cavity shape as calculated by Heywood circularity, (16) the standard deviation of cavity shape, (17) number of defects (counts), (18) image confidence (the area weighted average of confidence score). These 18 features were incorporated into our initial feature matrix. 

\begin{figure*}[ht]
 \centering
 \includegraphics[width=0.65\textwidth]{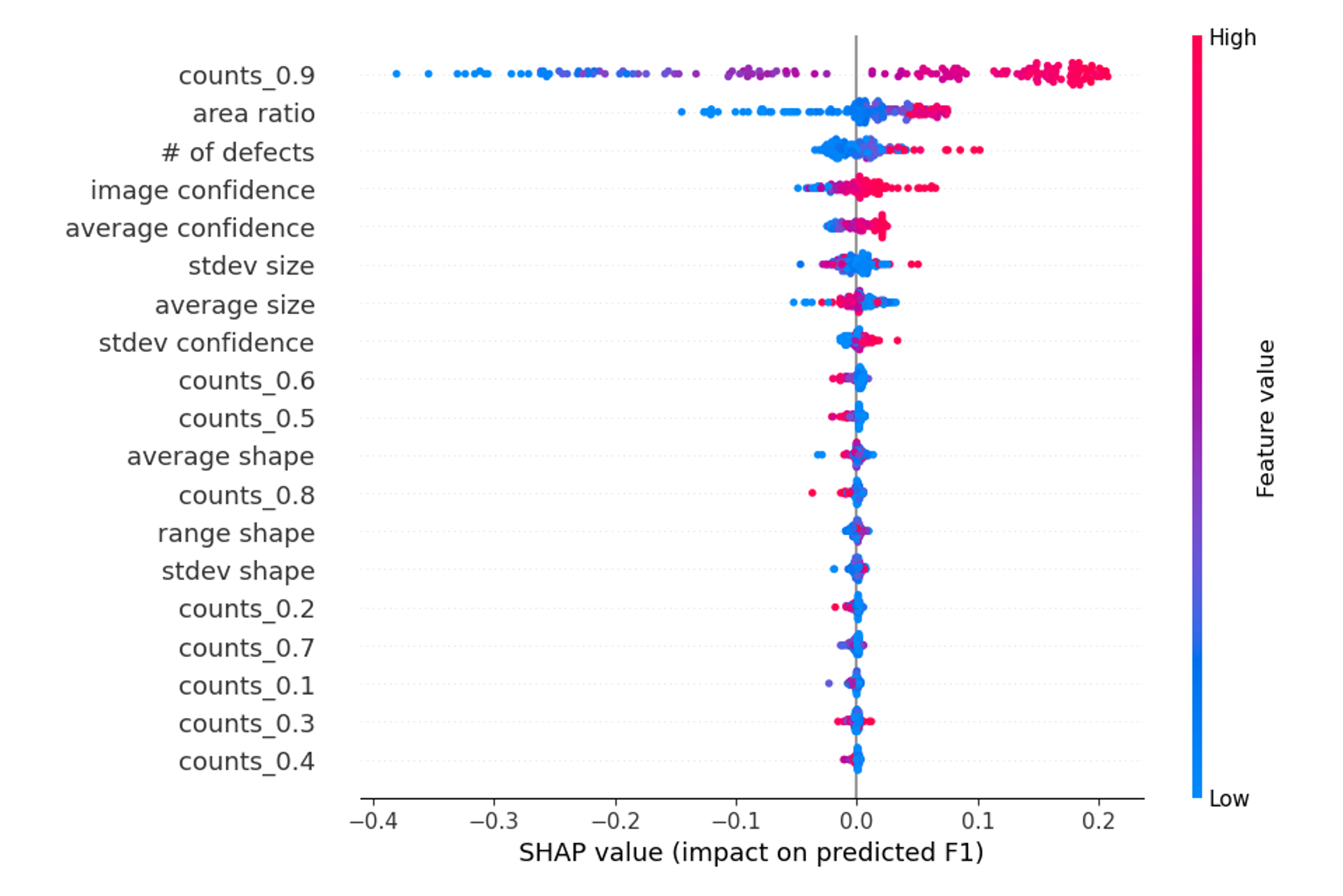}
\caption{SHAP value analysis of all feature candidates.}
 \label{fig:shap_all}
 \end{figure*}

We normalized all features to the same scale using the StandardScaler tool from the scikit-learn package to prevent any single feature from dominating the model due to its value range. To identify the most important features for our model, we conducted SHAP (SHapley Additive exPlanations) analysis \cite{NIPS20177062}. SHAP values provide a unified measure of feature importance by quantifying the contribution of each feature to the model's predictions. Figure \ref{fig:shap_all} presents the SHAP value summary plot for all feature candidates considered in the model, illustrating the impact of each feature on the model's output. SHAP values are used to interpret the contribution of each feature to the predictions. Each dot represents a SHAP value for a particular data point in the dataset, with colors indicating the feature value from low (blue) to high (red). The color gradient reveals how different values of the features affect the predictions. For instance, high values of the number of high-confidence defects (counts\_0.9) (red) tend to increase the SHAP value, positively influencing the model's output (i.e., high predicted F1 score), while low values (blue) have the opposite effect (i.e., low predicted F1 score). 

\begin{figure*}[ht]
 \centering
 \includegraphics[width=0.95\textwidth]{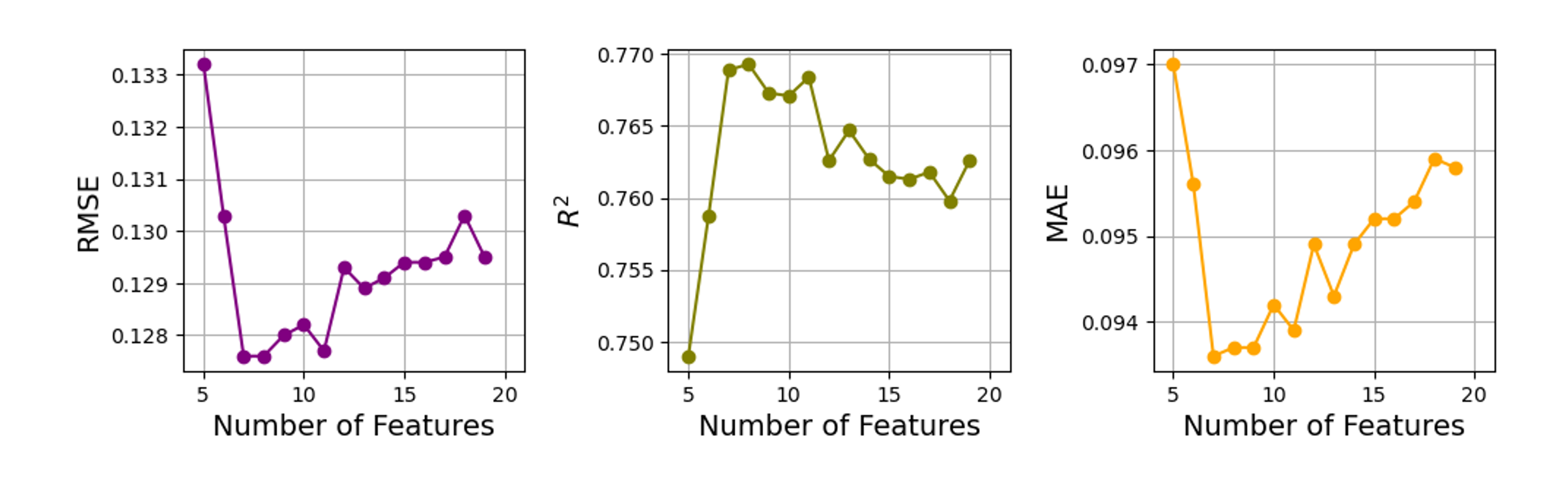}
\caption{RMSE, $R^2$, and MAE of the trained random forest model on test data as a function of the number of features used in the model.}
 \label{fig:feature_num}
 \end{figure*}
 
The features are listed on the y-axis, where feature (1-9) are denoted by "counts\_" followed by a number. For instance, counts\_0.1 represents the number of defects with confidence scores between 0.1 and 0.2. Feature (13) is denoted by "average size", and feature (14) was denoted by "std size". Based on the ranking from the SHAP analysis, we trained the random forest model using between 5 and 19 features. The resulting RMSE, $R^2$, and MAE are plotted as a function of the number of features as shown in Figure \ref{fig:feature_num}. The model achieved the best performance, with the lowest RMSE and highest $R^2$ score, when using the top eight features that had the most significant impact on the predictions. These eight features were therefore selected for the final model. Notably, the number of defects with confidence scores higher than 0.9 appears to have a greater impact on the model’s performance compared to the number of detected defects with lower confidence scores. This observation is reasonable because the number of high-confidence defects significantly influences both false positives and false negatives, thereby correlating strongly with the F1 score. Additionally, the average and standard deviation of the confidence score are crucial since they reflect the model’s ability to identify high-confidence detections reliably. Moreover, the average and standard deviation of fractional defect size are important factors; detecting small defects accurately poses a challenge for the model, influencing its overall performance. The area ratio and image confidence were also found to have a significant impact on the model’s output and were therefore adopted to train the random forest regression model.

\section{Results and Discussion}
\begin{figure*}[ht]
 \centering
 \includegraphics[width=0.95\textwidth]{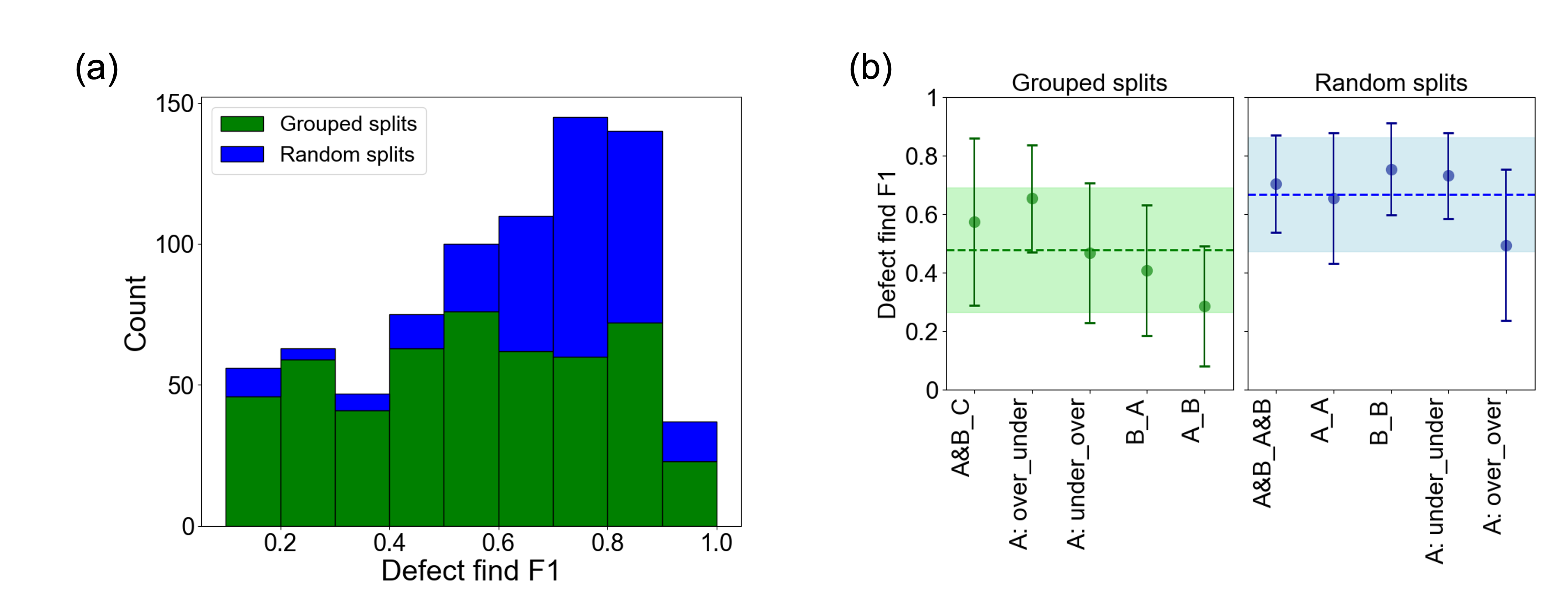}
\caption{(a) Histogram of defect find F1 scores. (b) The average defect find F1 scores with standard deviation error bars for different subsets of data. The dashed green and blue lines represent the average defect find F1 scores across all grouped and random split data, respectively. The green and blue shades depict the standard deviation over all data from grouped splits and random splits, respectively. Data labels indicate the different split of training and testing datasets.}
 \label{fig:data}
 \end{figure*}
The histogram shown in Figure \ref{fig:data} displays the distribution of all the Mask R-CNN defect find F1 scores of testing images obtained by evaluating the Mask R-CNN defect predictions against the ground truth labels on each image across different splits. The x-axis represents the range of F1 scores from 0 to 1, and the y-axis indicates the number of data points falling within each bin of F1 scores. It appears that the distribution of F1 scores from grouped splits is somewhat uniform since there are more than 80 instances falling within each bin. However, there are notably more instances with higher F1 score values for random splits. The mean F1 scores for various grouped splits and random splits are plotted in Figure \ref{fig:data}(b), with error bars representing the standard deviation for each split. Figure \ref{fig:data}(b) shows the random splits, where training and testing conditions are more likely to be drawn from the same distribution, have higher mean F1 scores. This result also serves as further evidence that the performance of the Mask R-CNN model depends on the similarity on the image domain between the training and testing datasets.

\begin{figure*}[ht]
 \centering
 \includegraphics[width=0.95\textwidth]{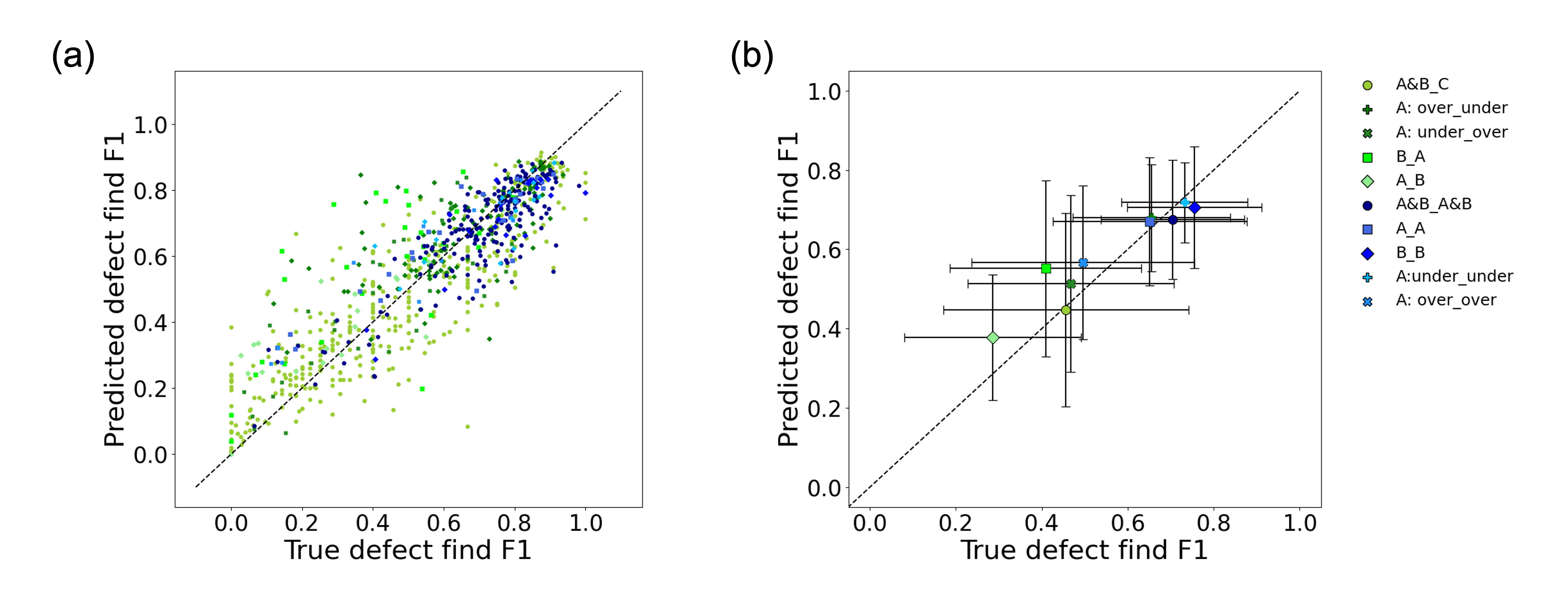}
\caption{(a) Parity plot comparing the predicted defect find F1-scores from the Random Forest model to the true scores across five test datasets from random five-fold cross-validation. Each symbol represents a different split within the datasets, and the details of the splits can be found in Table \ref{table:splits}. (b) Plot of mean predicted F1 scores for each split against mean true scores, with vertical  and horizontal error bars denoting standard deviation in predicted and true F1 scores, respectively. Dashed lines indicate the line of perfect prediction where predicted scores match true scores exactly.}
 \label{fig:parity}
 \end{figure*}

\begin{figure}[ht]
 \centering
 \includegraphics[width=0.475\textwidth]{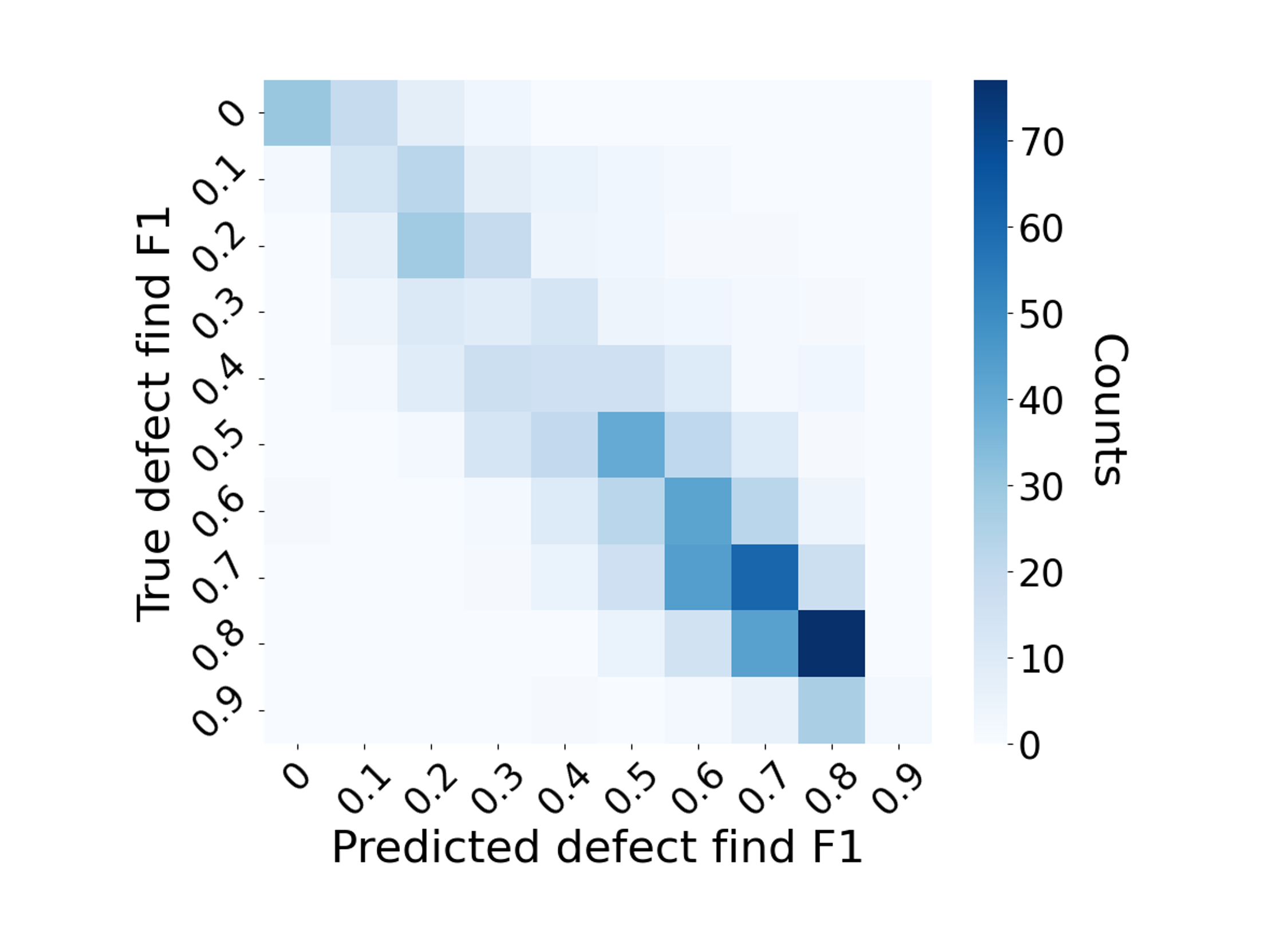}
\caption{Confusion matrix of the categorized F1 score prediction.}
 \label{fig:confu}
\end{figure}

The parity plot in Fig. \ref{fig:parity} (a) visualizes the performance of our random forest  model used to predict the defect find F1-score from random five-fold cross-validation. The dispersion of points along the line of parity (where the predicted score equals the true score) suggests a moderately strong correlation, supported by an MAE score of 0.094, an RMSE score of 0.127, and a $R^2$ score of 0.774. These metrics indicate a good level of accuracy in the model predictions across all the data. However, it is also observed that lower F1 scores tend to be overestimated, while higher F1 scores tend to be underestimated, which is a common behavior of regression models as they seek to minimize overall error and balance predictions around the mean.

We also observed that data points with true defect find F1 scores below 0.5 tend to deviate further from the parity line. Given that grouped splits generally have lower true F1 scores, we plot the average predicted defect find F1 score for each split against the average true F1 score in Fig. \ref{fig:parity} (b) to illustrate the overall performance across different splits. These averages show a strong alignment with the true scores, as evidenced by an MAE of 0.047, an RMSE of 0.062, and an $R^2$ of 0.831, which surpass the collective metrics across all data. The predictions on random splits align more closely with the true F1 scores than those on grouped splits, where the average MAE for random splits is 0.082 whereas the average MAE from the grouped splits is 0.121.

The F1 scores obtained from the test dataset and the corresponding predictions from the random forest model were categorized into intervals to construct a confusion matrix which is shown in Fig. \ref{fig:confu}. This confusion matrix helps in evaluating the accuracy of our model predictions across different score ranges. The matrix shows darker shades along the diagonal from the top left to the bottom right, indicating a higher concentration of instances where the predicted F1 scores align closely with the true F1 scores. Lighter shades off the diagonal reveal fewer occurrences, suggesting that most predictions fall within the correct range. 
\begin{table*}[h!]
\centering
\begin{tabular}{lcccccr}
\hline
\rowcolor{gray!20} 
\textbf{Data Subset} & \textbf{RMSE} & \textbf{MAE} & \textbf{$R^2$} & \textbf{NRMSE} & \textbf{NMAE} & \textbf{Number of images} \\
\hline
All data            & 0.127 & 0.093 & 0.774 & 0.475 & 0.167 & 833 \\
\hline

A\&B\_C      & 0.130 & 0.096 & 0.714 & 0.457 & 0.221 & 368 \\
\hline
A: over\_under    & 0.138 & 0.096 & -0.012 & 0.753 & 0.147 & 107 \\
\hline

A: under\_over    & 0.136 & 0.116 & 0.600 & 0.566 & 0.247 & 39 \\
\hline
B\_A            & 0.228 & 0.188 & -0.099 & 1.024 & 0.460 & 28 \\
\hline
A\_B & 0.147 & 0.108 & 0.247 & 0.715 & 0.377 & 19 \\
\hline

\rowcolor{LightBlue}
Average of grouped splits & 0.156 & 0.121 & 0.290 & 0.703 & 0.290 & 112.2 \\ \hline
A\&B\_A\&B & 0.092 & 0.069 & 0.652 & 0.550 & 0.098 & 195 \\ \hline
A\_A & 0.098 & 0.077 & 0.666 & 0.441 & 0.118 & 29 \\ \hline
B\_B & 0.100 & 0.077 & 0.645 & 0.639 & 0.102 & 19 \\ \hline
A: under\_under & 0.103 & 0.077 & -0.150 & 0.702 & 0.105 & 18 \\ \hline
A: over\_over & 0.141 & 0.110 & 0.409 & 0.543 & 0.223 & 11 \\ \hline
\rowcolor{LightGreen}
Average of random splits & 0.107 & 0.082 & 0.444 & 0.575 & 0.129 & 54.4 \\
\hline
\end{tabular}
\caption{Random forest regression model performance metrics across different data splits.}
\label{table:performance_metrics}
\end{table*}

Table \ref{table:performance_metrics} summarizes the model performance metrics obtained from both the grouped cross-validation and ten iterations of random five-fold cross-validation processes. The first row of the table summarizes metrics for the entire dataset, showing an RMSE of 0.127, an MAE of 0.093, and an $R^2$ score of 0.774 based on 833 test images. The next five rows provide metrics for grouped splits, ordered by the number of data points within each split. The average metrics over the five grouped splits are shown in the next row shaded in light blue. Similarly, metrics for random splits are shown in the following five rows, with the average over random splits displayed in the last row shaded in light green. RMSE and MAE vary the least across different data splits. In contrast, the $R^2$ score, NRMSE, and NMAE are influenced by the F1 score range within a split, often indicating higher errors for splits with narrower F1 score ranges. The average RMSE and MAE of the grouped splits are slightly higher than those for all data, while the average RMSE and MAE of the random splits are slightly lower than those for all data, suggesting higher prediction accuracy on randomly split data. An exception is observed in the split A:over\_over, which shows an RMSE of 0.141 and an MAE of 0.11, likely due to the limited number of just 11 data points and the low average F1 score in this split. 

\begin{figure}[h]
 \centering
 \includegraphics[width=0.475\textwidth]{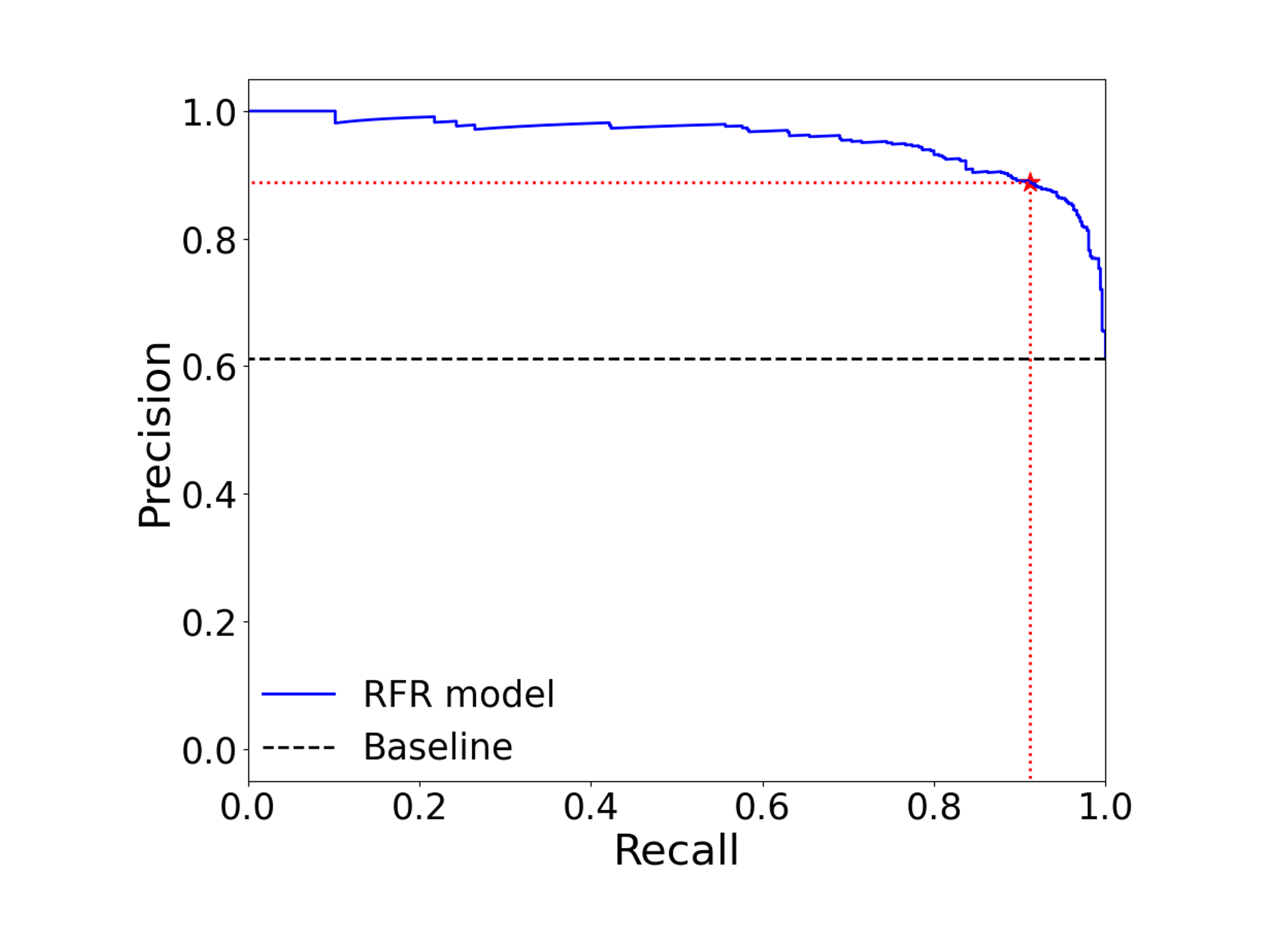}
\caption{Precision-Recall curve for domain estimation with threshold of 0.5 on defect find F1 score. The star marks the precision and recall at the selected F1 threshold of 0.5, which are 0.89 and 0.91, respectively.}
 \label{fig:pre-rec}
 \end{figure}
 
In the application context of the trained random forest model, one goal is to guide users in assessing if the results of defect detection on certain EM images using a trained Mask R-CNN model are reliable or not. This scenario can be framed as  a binary classification task. The F1 score predictions can be transformed into binary classifications by applying a threshold to the defect find F1 score. This precision-recall curve shown in Figure \ref{fig:pre-rec} illustrates the performance of the trained random forest model in classifying data points with a threshold of 0.5 on the defect find F1 score. We note that the choice of threshold is subjective, and for our present use-case the F1 threshold of 0.5 broadly divides reasonably well- vs. poor-performing images while simultaneously providing a robust ability of our random forest model to classify such well vs. poor-performing images. The solid blue line represents the precision of the random forest model at various thresholds of recall. The curve starts with a high precision close to 1.0 and gradually declines as recall increases, indicating that the model maintains a high precision across a wide range of recall levels before it begins to fall off. The dashed line represents the no-skill baseline, which indicates the performance of a model that would randomly guess the class. The performance of the random forest model is notably above this baseline, indicating its capability to discriminate between in- vs. out-of-domain (based on defect find F1 threshold of 0.5) effectively.

\begin{figure*}[h]
 \centering
 \includegraphics[width=0.95\textwidth]{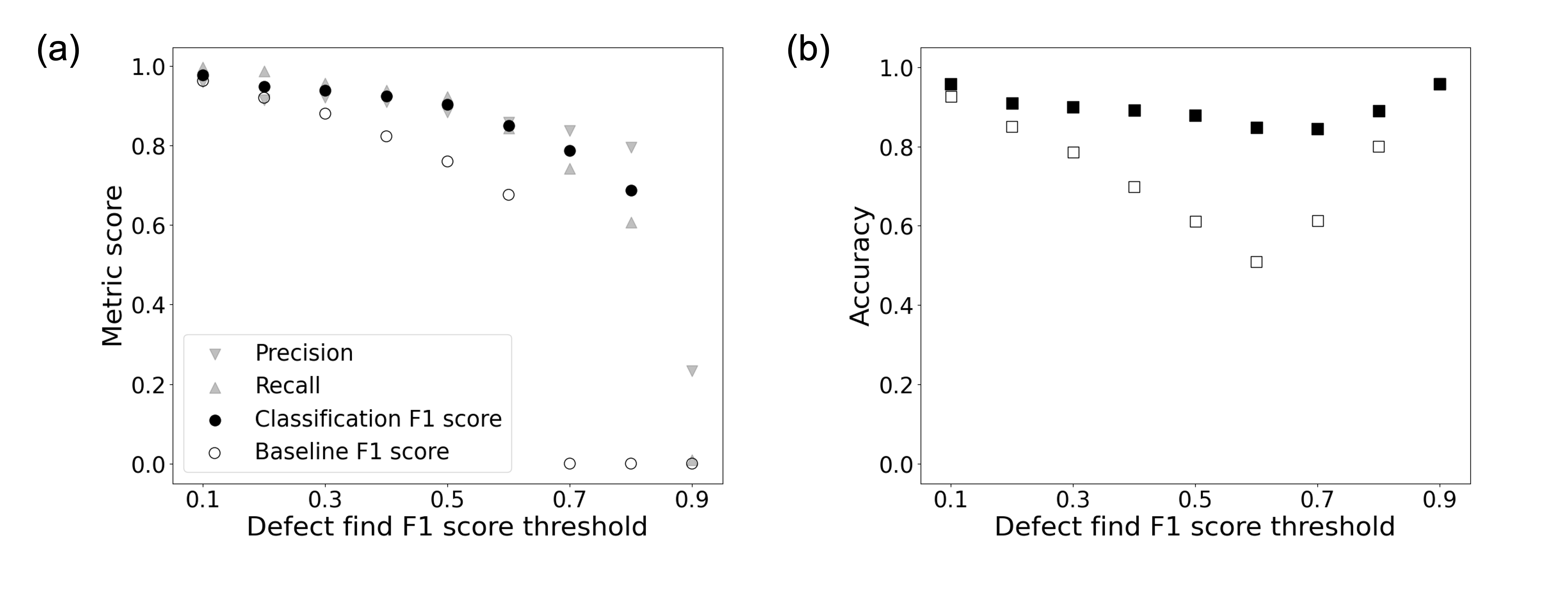}
\caption{(a) Domain classification metric scores: precision, recall and classification F1 score at various defect find F1 threshold. (b) Domain classification accuracy across various defect find F1 thresholds.}
 \label{fig:calssification}
 \end{figure*}

Figure \ref{fig:calssification} presents two plots comparing the performance of domain classification as a function of different defect find F1 score thresholds. The left plot illustrates the domain classification F1 score, and the right plot shows the domain classification accuracy (Acc), both as a function of various defect find F1 thresholds. In both plots, the solid colored dots represent the performance of the random forest model, while the lighter dots denote a baseline for comparison. Overall, the classification performance is significantly better than the baseline model, with a classification F1 score higher than 0.7 and classification accuracy exceeding 0.8 when the threshold on defect find F1 score is smaller than 0.8. As the threshold increases from 0.1 to 0.7, we also observe a general trend of decreasing domain classification F1 scores and accuracy.

In addition to evaluating the overall F1 score, we also trained random forest models to predict defect find precision and recall to gain a more nuanced understanding of our model’s performance. While the F1 score provides a balanced measure of both precision and recall, predicting these metrics independently allows us to assess specific aspects of the model’s capability. Precision indicates how many of the detected defects are true positives, highlighting the model's accuracy in defect identification. Recall, on the other hand, measures how many actual defects were detected, reflecting the model's ability to identify all relevant defects. 

Our model demonstrated strong performance in predicting precision, achieving MAE of 0.094, a RMSE of 0.132, and a $R^2$ score of 0.81. In contrast, predicting recall proved to be more challenging. The model for recall showed an MAE of 0.14, an RMSE of 0.192, and a $R^2$ score of 0.57. The evaluation metrics on predicting defect find precision, recall and F1 scores are summarized in Table \ref{table:f1-pre-recall}. Detailed analyses are provided in the SI. The model's performance in predicting precision surpasses that of predicting F1 score, as precision directly correlates with detected defects. However, predicting recall is more difficult because it involves estimating defects that the model failed to detect, which is inherently more challenging for machine learning models. 
\begin{table}[h!]
\centering
\begin{tabular}{lcccccr}
\hline
\textbf{Target} & \textbf{RMSE} & \textbf{MAE} & \textbf{R\textsuperscript{2}} & \textbf{NRMSE} & \textbf{NMAE} \\ \hline
\textbf{Precision}         & 0.132         & 0.094        & 0.81                         & 0.435          & 0.163         \\ \hline
\textbf{Recall}            & 0.192         & 0.140        & 0.57                         & 0.656          & 0.198         \\ \hline
\textbf{F1 score}          & 0.127         & 0.093        & 0.774                        & 0.475          & 0.167         \\ \hline
\end{tabular}
\caption{Performance metrics of random forest models on predicting precision, recall, and F1 score.}
\label{table:f1-pre-recall}
\end{table}

We also attempted to train a random forest model predicting swelling error of Mask R-CNN. However, the model shows poor performance with an $R^2$ score of 0.131. This outcome is expected, as predicting swelling error requires knowledge of the sizes of defects missed by the Mask R-CNN model. Without information about these undetected defects, estimating their sizes becomes significantly more challenging. Additional details can be found in the SI.

The Mask R-CNN model and the trained RF model using all the data we have is available on []. The trained Mask R-CNN model is designed specifically for detecting and segmenting cavity defects in TEM images, and thus, it is not intended for use with images outside this domain. To evaluate the usefulness and reliability of the random forest model, we tested it on COCO-128 images\cite{lin2015microsoftcococommonobjects}, which significantly differ from EM images. We observed that Mask R-CNN often over-confidently detected cavities in these images, despite the absence of any actual cavities, resulting in an expected F1 score of 0. The random forest model, however, produced predicted defect F1 scores below 0.7, with more than 75\% of them falling below 0.5. Examples of Mask R-CNN output images and the histogram of predicted F1 scores from the random forest model are provided in the SI. Although these predictions are not close to 0, they are still substantially lower than those for EM images in random splits. This contrast, with the Mask R-CNN’s overconfidence and the moderate F1 scores of the random forest, suggests that the random forest model successfully captures features indicative of domain estimation, showing potential for identifying out-of-domain images.

\section*{Summary and Conclusion}

Our study presents a flexible and practical approach to assess the accuracy of an object detection model on new images, particularly when ground truth labels are unavailable. The approach uses a random forest regression model to learn the F1 score of the underlying object detection model based on features from the model detections and confidence scores, allowing F1 to be predicted for new images processed by the object detection model. We demonstrate our approach using Mask R-CNN models trained to detect cavities in TEM images of irradiated metal alloys. The random forest regression model's predictions of the defect detection F1 score closely mirror the true performance, as evidenced by the MAE of 0.093, $R^2$ score of 0.774, and the high concentration of accurate predictions in the confusion matrix. The robustness of our method was validated across various splits of data, though the performance on splits grouped by different image characteristics is relatively worse than on random splits.

By enabling users to predict model performance on new, unlabeled data, we bridge a significant gap in automated defect detection workflows. In particular, the approach taken here could be used to provide automatic guardrails for users of defect detection models, warning them when prediction quality is a concern. Moreover, the success of this methodology paves the way for future research to extend such performance estimation to other deep learning models in materials science and beyond. 

\section*{Author contributions}
N.L. preprocessed data from Mask R-CNN, built and trained the random forest model, conducted feature engineering, and wrote the manuscript.
R.J. trained and tested the Mask R-CNN model, generated outputs for the random forest model, contributed ideas for useful features, and revised the manuscript.
M.L. and K.F. contributed to discussions and revisions.
V.A. contributed to discussions and explored alternative approaches to address the problem.
D.M. contributed to discussions, provided many ideas, revised the manuscript, and serves as the corresponding author.
\section*{Conflicts of interest}
There are no conflicts to declare.

\section*{Data and code availability}
The data used to train the random forest model is available on Figshare\url{https://doi.org/10.6084/m9.figshare.27281400.v1}. The code for evaluating the trained Mask R-CNN model, extracting features from the Mask R-CNN output and for training the random forest model are available on Github \url{https://github.com/uw-cmg/cavity_defect_detection/tree/main}. 
\section*{Acknowledgments}

This research was supported by the Electric Power Research Institute (EPRI) under award number 10012138. We utilized the Extreme Science and Engineering Discovery Environment (XSEDE), funded by National Science Foundation Grant ACI-1548562. The training and testing of the Mask R-CNN model were conducted on the Bridges-2 system through allocation TG-DMR090023, funded by NSF award ACI-1928147, at the Pittsburgh Supercomputing Center (PSC).
%Bibliography
\bibliographystyle{plainnat}
\bibliography{arxiv}  

\end{document}